\documentclass[11pt]{article}
\usepackage{multirow}

\usepackage[preprint]{acl}

\usepackage{times}
\usepackage{latexsym}
\usepackage[T1]{fontenc}

\usepackage{booktabs}
\usepackage[utf8]{inputenc}

\usepackage{microtype}

\usepackage{inconsolata}

\usepackage{graphicx}
\usepackage{listings}
\title{MASFactory: A Graph-Centric Framework for Orchestrating LLM-Based Multi-Agent Systems with Vibe Graphing}

\author{
  \textbf{Yang Liu\textsuperscript{1}}, \textbf{Jinxuan Cai\textsuperscript{1}}, \textbf{Yishen Li\textsuperscript{1}}, \textbf{Qi Meng\textsuperscript{1}}, \textbf{Zedi Liu\textsuperscript{1}}, \\ 
  \textbf{Xin Li\textsuperscript{1}}, \textbf{Chen Qian\textsuperscript{2}}, \textbf{Chuan Shi\textsuperscript{1}} \and\textbf{Cheng Yang\textsuperscript{1}\thanks{\ \ Corresponding author.}} \\
  \textsuperscript{1}Beijing University of Posts and Telecommunications \\
  \textsuperscript{2}Shanghai Jiao Tong University \\
  \texttt{\{liuyang1999, yangcheng\}@bupt.edu.cn}
}

\begin{document}
\maketitle

\begin{abstract}
Large language model-based (LLM-based) multi-agent systems (MAS) are increasingly used to extend agentic problem solving via role specialization and collaboration. MAS workflows can be naturally modeled as directed computation graphs, where nodes execute agents/sub-workflows and edges encode dependencies and message passing.
However, implementing complex graph workflows in current frameworks still requires substantial manual effort, offers limited reuse, and makes it difficult to integrate heterogeneous external context sources.
To overcome these limitations, we present MASFactory, a graph-centric framework for orchestrating LLM-based MAS.
It introduces Vibe Graphing, a human-in-the-loop approach that compiles natural-language intent into an editable workflow specification and then into an executable graph.
In addition, the framework provides reusable components, skill support, multimodal message handling, and pluggable context integration, as well as a visualizer for topology preview, runtime tracing, and human-in-the-loop interaction. 
We evaluate MASFactory on seven public benchmarks, validating both reproduction consistency for representative MAS methods and the effectiveness of Vibe Graphing.
Our code\footnote{\url{https://github.com/BUPT-GAMMA/MASFactory}. Licensed under Apache-2.0; use, modification, and distribution are permitted within its terms.} and video\footnote{\url{https://youtu.be/ANynzVfY32k}} are publicly available.
\end{abstract}

\section{Introduction}
\begin{figure}[t]
    \centering
    \includegraphics[width=\linewidth, height=11cm, keepaspectratio]{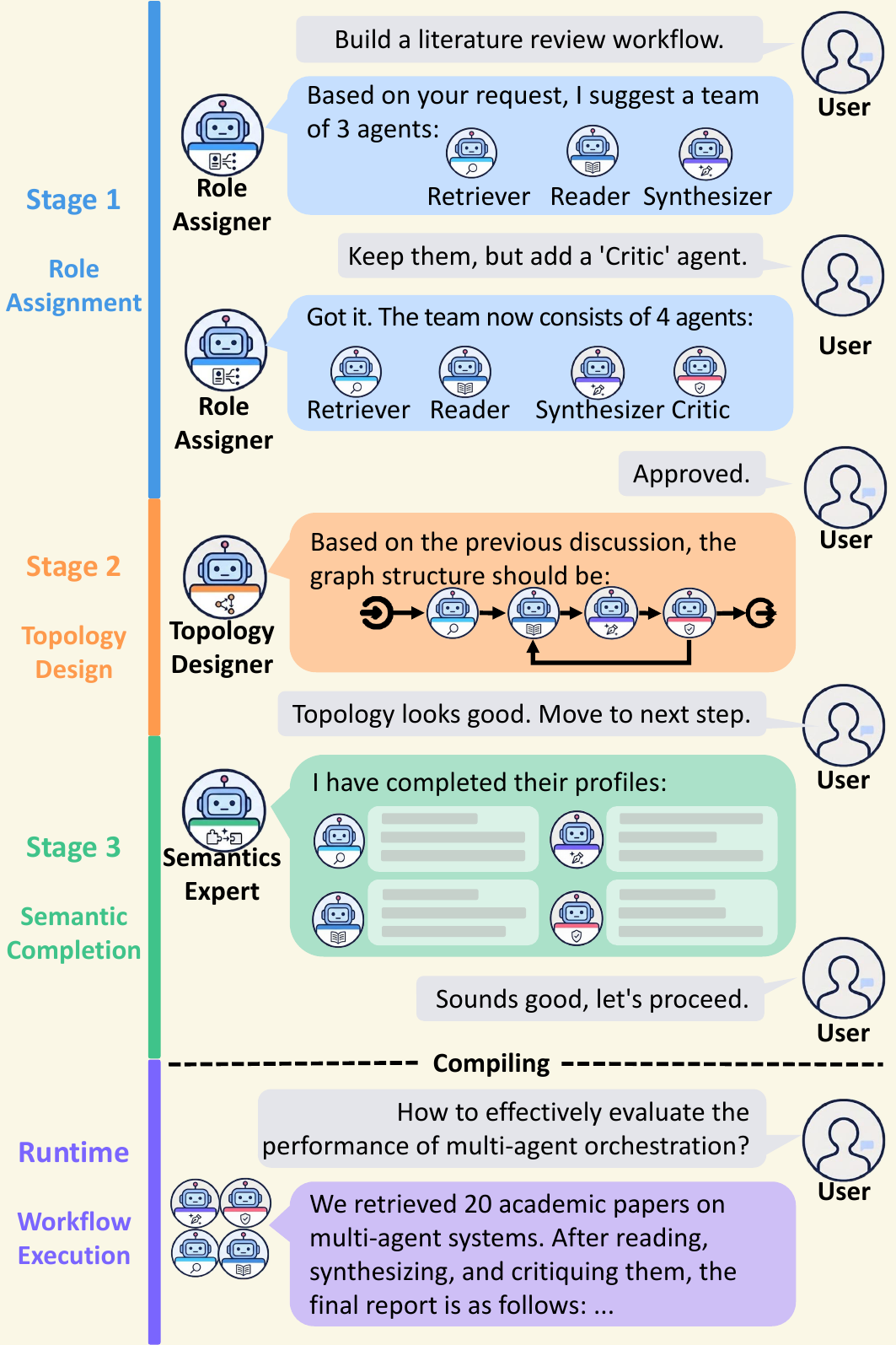}
    \caption{
    \textbf{Vibe Graphing in MASFactory.}
    MASFactory turns a user's natural-language intent into an executable multi-agent workflow via a three-stage, human-in-the-loop process, then compiles and executes the resulting workflow at runtime.}
    \label{fig:vibegraphing}
\end{figure}

With the introduction of tool use and feedback-driven correction, large language models (LLMs) are often packaged as agents that follow a \emph{perception--reasoning--action} loop, enabling long-horizon task execution in external environments~\citep{yao2023react,schick2023toolformer,shinn2023reflexion}.
As tasks grow in complexity, a single agent often struggles to simultaneously achieve end-to-end coverage, robust error recovery, and multi-step coordination. Consequently, LLM-based multi-agent systems (MAS) have become a widely adopted approach to extend agentic problem solving through role specialization, cross-checking, and iterative collaboration ~\citep{li2023camel,wu2023autogen,hong2024metagpt,qian2024chatdev,chen2024agentverse}.

For MAS modeling, an increasingly common orchestration abstraction is directed computation graphs.
A graph consists of nodes that host computation units (agents, tools, or sub-workflows) and edges that encode execution dependencies and message-passing directions.
Pure directed acyclic graph (DAG) workflows are well-suited for pipeline-style collaboration, while cyclic structures are crucial for iterative processes such as reflection, revision, and retry.
Recent frameworks have begun to explicitly represent multi-agent workflows as graphs. For instance, LangGraph models workflow as a \emph{stateful graph} with explicit execution semantics \citep{langgraph}, and Dify provides a workflow canvas where DAG dependencies define both execution order and data flow \citep{dify}.

However, implementing complex MAS remains engineering-intensive. 
For example, developers have to manually craft role prompts for agent nodes, wire the routing logic between nodes, and establish inter-agent communication protocols. 
Also, real-world applications rely on heterogeneous context sources, including reusable skill packages for task-specific guidance and auxiliary resources~\citep{anthropic_skills}, memory layers~\cite{chhikara2025mem0,memoryos2025,packer2023memgpt}, retrieval-augmented generation (RAG)~\cite{lewis2020rag,graphrag}, and standardized tool/context integration protocols such as the model context protocol (MCP)~\cite{mcp}. Current frameworks often integrate them via workflow-specific glue code, making implementation difficult to port and reuse across environments. Finally, MAS development frequently involves repeated subgraphs that are globally similar but differ slightly in local configurations. Existing frameworks provide limited support for version-controlled, templated reuse of such subgraphs.

To address these challenges, we present \textbf{MASFactory}, a composable orchestration framework for multi-agent systems with \textbf{Vibe Graphing} (illustrated in Figure~\ref{fig:vibegraphing}).
Firstly, MASFactory introduces Vibe Graphing: users describe design intent in natural language, and the system compiles it into a readable, editable, and version-controlled structured intermediate representation, which is further compiled into an executable workflow.
This process incorporates \emph{human-in-the-loop} interaction, allowing users to review, modify, and provide feedback through a visual interface.
Secondly, MASFactory improves interoperability via a Context Adapter that hides heterogeneity across different context sources, and further supports multimodal message handling by representing image and PDF inputs as structured media assets.
Thirdly, MASFactory improves reusability by providing configurable, reusable graph modules for common multi-agent collaboration patterns, enabling developers to quickly assemble workflows without repeatedly implementing structurally similar subgraphs.
Beyond these, MASFactory provides a visualizer that supports visual inspection of workflow topology, runtime trace visualization, and human-in-the-loop interaction.

\paragraph{Contributions.}
(1) We propose \textbf{MASFactory}, a graph-centric MAS orchestration framework with reusable components, skill support, multimodal message handling, pluggable context management, delivering both natural-language workflow generation and consistent reproducible results. (2) We introduce \textbf{Vibe Graphing}, a human-in-the-loop approach that compiles natural-language intent into executable graphs, reducing implementation effort while achieving competitive performance with manually implemented workflows. (3) We reproduce five representative MASs with MASFactory, and demonstrate competitive performance on seven public benchmarks.

\section{Related Work}

\paragraph{LLM-based Multi-Agent Systems.}
Recent work has demonstrated that coordinating multiple LLM agents can improve robustness and coverage via role specialization, cross-checking, and iterative collaboration~\cite{li2023camel,chen2024agentverse,hugginggpt2023}.
For example, AutoGen popularizes programmable multi-agent conversation patterns and group coordination \citep{wu2023autogen}.
MetaGPT structures collaboration with role definitions and SOP-like procedures for complex tasks \citep{hong2024metagpt}. ChatDev organizes software-development collaboration into staged role interactions \citep{qian2024chatdev}. However, the implementations of these works vary significantly, and their codebases often span from thousands to tens of thousands of lines, making it difficult for developers to build upon them.

\paragraph{Frameworks for MAS Development.}
Beyond research prototypes, a growing ecosystem targets practical MAS implementation. Google's Agent Development Kit (ADK)~\cite{googleadk} and CrewAI~\cite{crewai} emphasize a code-first, process-driven methodology: the former provides a toolkit towards deploying agentic applications, whereas the latter focuses on the programmatic orchestration of collaborative agent crews. Alternatively, graph-centric frameworks such as LangGraph~\cite{langgraph} and Dify~\citep{dify} represent agentic workflows as explicit node/edge structures to provide fine-grained control over the execution flow. However, building complex MAS remains engineering-intensive, requiring heavy manual configuration and tightly coupled context integrations. 

To simplify orchestration, MASFactory complements graph-centric frameworks such as LangGraph and Dify by focusing on reusable and intent-driven MAS construction.
Unlike frameworks that mainly emphasize execution control or manual workflow editing, MASFactory represents workflows as editable graph specifications that can be generated via Vibe Graphing, inspected through the visualizer, and instantiated through reusable components.
This reduces the effort of constructing, adapting, and reusing MAS workflows.

\section{System Design}

\begin{figure*}[t]
    \centering
    \includegraphics[width=\textwidth]{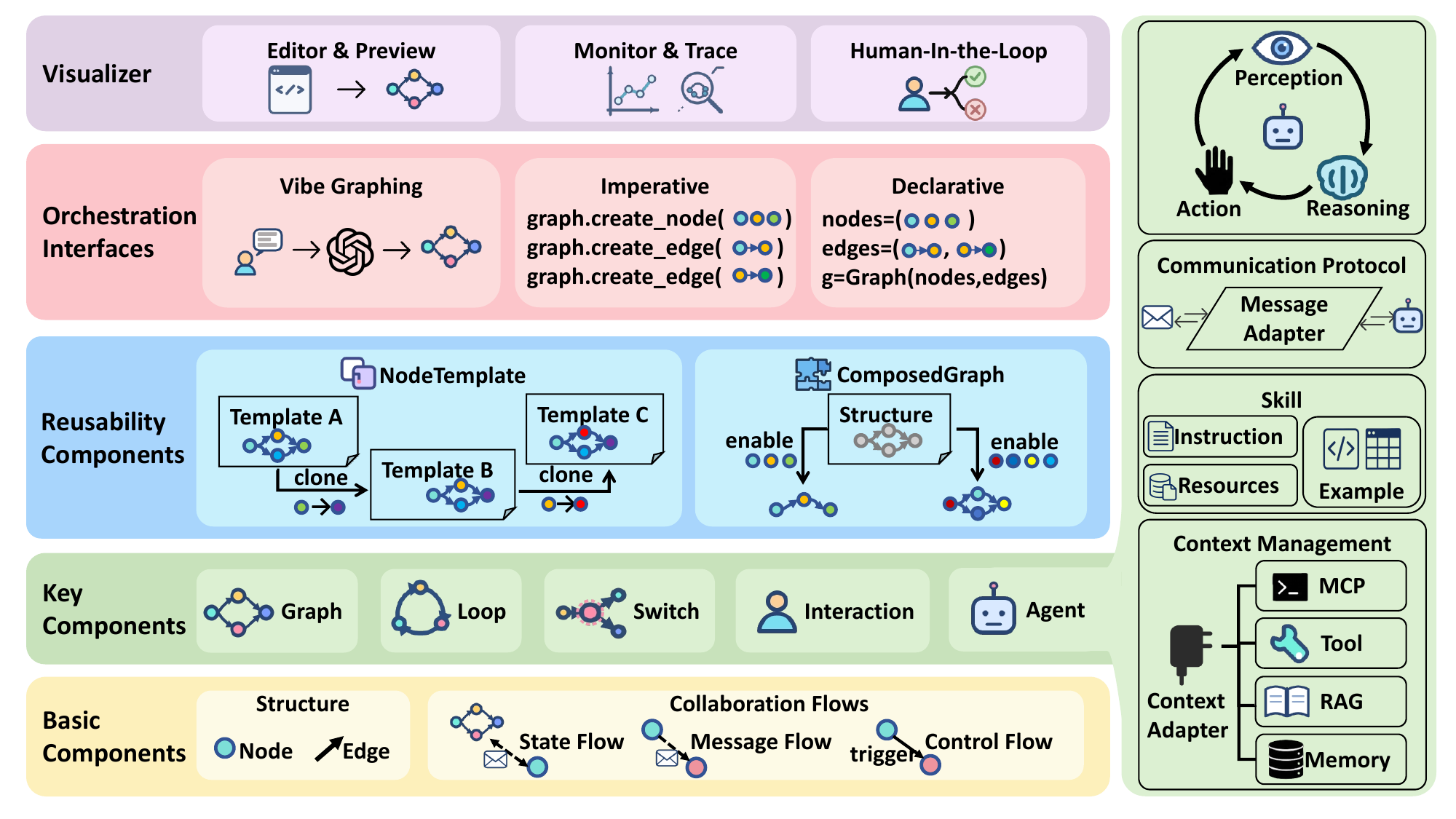}
    \caption{Architecture overview of MASFactory framework.}
    \label{fig:system_design}
\end{figure*}

In this section, we present the architecture of MASFactory.
As shown in Figure~\ref{fig:system_design}, at the bottom layer the system adopts a backbone composed of \texttt{Node} and \texttt{Edge} for modeling collaboration graphs.
Built on top of this backbone, MASFactory provides composable components and modules for flexible collaboration process implementation. 
Besides, MASFactory unifies MAS-critical mechanisms, including communication protocols, skill packaging, multimodal input handling, and context management, and builds them as pluggable modules.
As a result, the system can conveniently integrate external frameworks such as Mem0~\cite{chhikara2025mem0} and LlamaIndex~\cite{llamaindex} for better user experience. 
MASFactory also provides three orchestration interfaces: automated orchestration via Vibe Graphing, and manual workflow implementation via declarative or imperative programming. To facilitate development and debugging, MASFactory also provides a visualizer for topology preview, runtime tracing, and human-in-the-loop interactions throughout the workflow.

\subsection{Basic Components}
\textbf{Graph-based Organization.} MASFactory models collaboration as a directed graph composed of \texttt{Node} and \texttt{Edge} \citep{zhuge2024gptswarm}.
A \texttt{Node} is the basic computation unit and can be extended into freely composable components, including \texttt{Graph}, \texttt{Loop}, \texttt{Agent}, \texttt{CustomNode}, \texttt{Interaction}, and \texttt{Switch}.
These components expose a consistent interface, but differ in their execution logic.
An \texttt{Edge} expresses inter-node dependencies and serves as the carrier of message passing.

\textbf{Collaboration Flow.} 
MASFactory makes collaboration signals explicit by separating them into three flows.
\textbf{Control flow} propagates along edges to advance scheduling and dependencies, ensuring causal constraints of execution.
\textbf{Message flow} propagates horizontally along edges to carry node outputs to downstream nodes.
\textbf{State flow} propagates along the hierarchy between graphs and subgraphs to synchronize graph-level context and runtime state.

\textbf{Multimodal Message Representation.}
MASFactory extends message flow beyond plain text by representing images, PDFs, and textual content as structured message fields with explicit modality declarations.
This design allows multimodal inputs and intermediate outputs to be propagated through the same graph-level message-passing mechanism, while leaving model-specific serialization to downstream adapters.

\textbf{Node Lifecycle.}  At runtime, each node follows a unified lifecycle.
A node first aggregates incoming messages along its \texttt{in edges} into its input, then reads states from its parent \texttt{Graph}.
It subsequently runs node-specific execution logic, e.g., invoking the LLM for \texttt{Agent}, scheduling internal nodes in topological order for \texttt{Graph} and \texttt{Loop}.
After execution, the node dispatches produced messages via its \texttt{out edges} to downstream nodes, and finally writes updated states back to the parent \texttt{Graph} to synchronize shared states and support downstream execution.

\textbf{Runtime Scheduling.} MASFactory adopts a readiness-based scheduling strategy that allows multiple ready nodes to execute concurrently.
This unified mechanism supports sequential, parallel, branching, and cyclic control structures.

\subsection{Key Components}
\textbf{Graph and Loop.} \texttt{Graph} and \texttt{Loop} are responsible for topology representation and scheduling of internal nodes.
\texttt{Graph} expresses and schedules directed acyclic graph (DAG) workflows.
\texttt{Loop} expresses and schedules cyclic structures, and is commonly used for iterative collaboration patterns such as reflection, revision, and retry.

\textbf{Switch.} A \texttt{Switch} implements control-flow routing inside the graph.
Unlike ordinary nodes that broadcast messages and trigger signals to all downstream nodes by default, \texttt{Switch} dynamically selects and activates one or multiple downstream paths based on runtime state.

\textbf{Interaction.} An \texttt{Interaction} node serves as the entry point of the human-in-the-loop mechanism.
It can actively query users during execution, collect feedback, and inject user inputs back into the workflow.
\subsection{Agent Component}
\textbf{Perception--Reasoning--Action.} \texttt{Agent} adopts the classic Perception--Reasoning--Action paradigm and modularizes key steps such as communication and context through pluggable designs.
As indicated in Figure~\ref{fig:system_design}, \texttt{Agent} relies on a pluggable \emph{Message Adapter} for message processing and a pluggable \emph{Context Adapter} for context management.

\textbf{Message Adapter.} For communication, the \texttt{Message Adapter} formats agent inputs and outputs according to a given communication protocol. 

MASFactory provides a set of commonly used adapters, including protocols based on JSON schema, structured Markdown segments, and plain-text paragraph formats.
It also exposes interfaces for user-defined protocols.
By introducing Message Adapters, MASFactory decouples collaboration graph from protocol definitions, enabling protocol extension or replacement without modifying the collaboration topology.

\textbf{Context Adapter.} For context management, to shield heterogeneity across information sources such as Memory, MCP, and RAG \citep{packer2023memgpt,memoryos2025,mcp,lewis2020rag,graphrag}, MASFactory utilizes a Context Adapter to provide a standardized interface. This adapter segments diverse external contexts into standardized units, thereby enabling unified seamless integration with different context frameworks such as Mem0 and LlamaIndex.

\textbf{Skill Support.}
Inspired by recent Agent Skills practice, MASFactory supports lightweight skill packages that encapsulate reusable task guidance and auxiliary resources, allowing agents to reuse task-specific capabilities without changing the workflow topology.

\subsection{Reusability and Templated Prototyping}
\textbf{NodeTemplate.} NodeTemplate allows users to declare a structural template first and instantiate it into a concrete graph later.
This decoupling between declaration and instantiation enables users to clone templates to build multiple graphs that share a similar global structure but differ in local configurations, supporting branch-style reuse and versioned management.
Moreover, developers can reuse a graph-level template while adjusting node-level settings when instantiating the graph.

\textbf{ComposedGraph.} ComposedGraph is a specialized form of \texttt{Graph} that represents a class of predefined structures.
It can instantiate a concrete graph by filling node configurations or activating specific branches according to user parameters, hiding low-level construction details.
With ComposedGraph, users can package their designs as reusable composite components.
Meanwhile, MASFactory uses ComposedGraph to encapsulate structures that are difficult to express with static graphs, such as DyLan-style dynamic scheduling patterns \citep{liu2023dylan}, as well as commonly used collaboration subgraphs.
This supports broader application scenarios while reducing development cost.

\subsection{Orchestration Interfaces}
\textbf{Vibe Graphing.} Under Vibe Graphing, MASFactory compiles natural-language intent into an executable multi-agent workflow through staged compilation, while providing human-in-the-loop review and revision at each stage.
During this process, the system produces a readable, editable and structured intermediate representation, which is then compiled into an executable workflow.
As illustrated in Figure~\ref{fig:vibegraphing}, the pipeline completes three core tasks.
\emph{Role Assignment} maps task intent into a set of candidate agents with clear responsibility boundaries.
\emph{Structure Design} generates a directed-graph topology skeleton based on inter-role information dependencies and control constraints, determining connectivity as well as the directions of message and control propagation.
\emph{Semantic Completion} performs parameterized instantiation over the skeleton by configuring prompts and tools for each node, producing a workflow that can be compiled and executed directly.
This intent--structure--instantiation compilation chain elevates MAS construction from manual workflow configuration to an iterative design process, reducing implementation cost while retaining control over topology and semantics.

\textbf{Imperative Interface.} The imperative interface is code-centric: developers construct graphs by programmatically instantiating nodes and edges and wiring them with explicit control logic. This style offers high flexibility and precise control over topology, parameters, and runtime behaviors, making it suitable for carefully engineered workflows and scenarios that require tight coupling with application logic.

\textbf{Declarative Interface.} The declarative interface specifies a workflow as a structured configuration: developers declare the graph topology and node properties, and MASFactory constructs the executable graph accordingly. This style keeps programming concise and easy to review, and is recommended for fixed workflows and lightly dynamic structures.

\subsection{Visualizer}
The Visualizer is a visual integrated environment implemented as a VS Code extension.
It aligns static workflow topology with runtime traces in a single view, and supports the following features:

\textbf{Editor \& Preview.} 
Editor \& Preview supports real-time topology preview and structural inspection during development.

\textbf{Monitor \& Trace.} The visualizer tracks node state evolution and message propagation during execution to support debugging and diagnosis.

\textbf{Human-in-the-Loop.} Human-in-the-Loop works with \texttt{Interaction} nodes to visualize runtime user interactions and to incorporate external feedback or inputs into the Vibe Graphing workflow, enabling interactive intervention and iteration.

\section{Evaluation and Analysis}
\label{sec:eval}
\begin{table*}[t]

\centering
\small
\renewcommand{\arraystretch}{1.15} 
\setlength{\tabcolsep}{4pt}
\caption{Main results are reported on a percentage scale (0-100). The symbol ``--'' indicates not applicable due to the incompatibility between programming-focused MASs and general-purpose reasoning benchmarks.}
\label{tab:main_results}
\begin{tabular}{lccccccc}

\toprule
\textbf{Method} &
\textbf{HumanEval} &
\textbf{MBPP} &
\textbf{BigCodeBench} &
\textbf{SRDD} &
\textbf{MMLU-Pro} &
\textbf{GAIA} &
\textbf{GPQA}
\\
\midrule
ChatDev (original)                  & 82.50 & 71.40 & 50.70 & 82.91 & --    & --    & -- \\
ChatDev (MASFactory)            & 81.30 & 74.20 & 53.30 & 84.23 & --    & --    & -- \\
\midrule
MetaGPT (original)                  & 67.07 & 36.03 & 50.10 & 78.19 & --    & --    & -- \\
MetaGPT (MASFactory)            & 89.02 & 59.14 & 51.70 & 72.77 & --    & --    & -- \\
\midrule
AgentVerse (original)               & 85.00 & 74.54 & 65.92 & 87.55 & 64.64 & 12.12 & 38.39 \\
AgentVerse (MASFactory)         & 85.00 & 75.15 & 64.12 & 91.06 & 64.16 & 12.73 & 37.50 \\
\midrule
CAMEL (original)                    & 62.20 & 60.60 & 63.51 & 89.42 & 50.08 & 9.70  & 32.59 \\
CAMEL (MASFactory)              & 71.85 & 57.80 & 78.16 & 89.69 & 63.04 & 12.73 & 24.78 \\
\midrule
HuggingGPT (original)              & 82.32 & 68.60 & 28.42 & 87.96 & 65.59 & 9.09  & 56.67 \\
HuggingGPT (MASFactory)        & 80.49 & 64.40 & 29.91 & 83.26 & 63.66 & 10.91 & 47.32 \\
\midrule
Vibe Graphing-ChatDev        & 83.50 & 74.20 & 45.30 & 88.13 & --    & --    & -- \\
Vibe Graphing-Task Specific                 & 84.76 & 72.37 & 51.67 & 90.71 & 51.73 & 12.12 & 39.51 \\
\bottomrule
\end{tabular}

\end{table*}

We conduct extensive experiments to answer the following questions: (1) whether MASFactory can reproduce representative MAS methods with consistent effectiveness; (2) whether Vibe Graphing can produce competitive workflows compared with manually designed ones; and (3) whether MASFactory can reduce implementation cost through component reuse and intent-driven orchestration.
\subsection{Experimental Setup}
\textbf{Benchmarks.}
We evaluate on coding-oriented benchmarks (HumanEval, MBPP, BigCodeBench, SRDD) \citep{chen2021codex,mbpp,bigcodebench2024,qian2024chatdev},
and general reasoning/tool-use benchmarks (MMLU-Pro, GAIA, GPQA) \citep{mmlupro2024,gaia2023,gpqa2023}.
All scores are reported in percentage scale (0--100).
Coding benchmarks except SRDD report pass@1; SRDD reports an aggregate quality score (SRDD);
MMLU-Pro/GAIA/GPQA report accuracy.

\textbf{LLM Backbones.} For Vibe Graphing, the \emph{workflow construction} stage uses \texttt{gpt-5.2} to compile natural-language intent into an executable workflow. For workflow execution of all methods, we use \texttt{gpt-4o-mini}.

\subsection{Performance of Reproduced Workflows}
Table~\ref{tab:main_results} compares five representative MAS methods implemented in MASFactory against their original implementations,
including ChatDev \citep{qian2024chatdev}, MetaGPT \citep{hong2024metagpt}, AgentVerse \citep{chen2024agentverse}, CAMEL \citep{li2023camel}, and HuggingGPT \citep{hugginggpt2023}.
Overall, MASFactory reproductions achieve results broadly consistent or even better with the originals across different benchmarks. This indicates that MASFactory can cover diverse multi-agent architectures and collaboration designs without systematic regressions.

\subsection{Performance of Vibe Graphing}
We report two Vibe Graphing settings in Table~\ref{tab:main_results}. 
In \emph{Vibe Graphing-ChatDev}, we replace each key phase of ChatDev with an individual \texttt{VibeGraph} component, and connect these phases into a workflow using lightweight glue code. 
In contrast, \emph{Vibe Graphing-Task Specific} is a task-driven variant: for each benchmark dataset, a developer writes a natural language description of a workflow, and passes the instruction to a single \texttt{VibeGraph} component to compile it into an executable workflow. The last two rows of Table~\ref{tab:main_results} reports the results for workflows produced with Vibe Graphing.
Despite using a different model only at the workflow construction stage (\texttt{gpt-5.2}), the resulting workflows execute with \texttt{gpt-4o-mini} and achieve competitive performance on coding benchmarks, as well as on general reasoning/tool-use tasks where applicable.
These results suggest that staged intent-to-graph compilation can generate viable multi-agent workflows that approach manually designed baselines, while substantially reducing low-level graph wiring and framework-specific engineering effort.
\subsection{Case Study of Implementation Cost}
\label{sec:Implementation Cost}

The original implementation of ChatDev contains 1{,}511 lines of Python code for workflow definition. 
With ComposedGraph-based reuse, our MASFactory reproduction reduces the total implementation to 1{,}114 lines while maintaining comparable performance as shown in Table~\ref{tab:main_results}. If we further use Vibe Graphing to make each ChatDev stage interactively generated (i.e., Vibe Graphing-ChatDev), the implementation only requires 203 lines to connect these stages. When we fully rely on Vibe Graphing to generate the workflow end-to-end (i.e., Vibe Graphing-Task Specific), the workflow specification shrinks to only 45 lines of code.

\begin{table}[!t]
\centering
\renewcommand{\arraystretch}{1.15}
\caption{Comparison of monetary cost between Vibe Graphing (VG) and Vibe Coding (VC). VC-L and VC-M denote the low and medium reasoning lengths of VC.}
\label{tab:vibe_cost}
\resizebox{\columnwidth}{!}{
\begin{tabular}{lcccccc}
\toprule
\multirow{2}{*}{\textbf{Metric}} & \multicolumn{3}{c}{\textbf{ChatDev}} & \multicolumn{3}{c}{\textbf{AgentVerse}} \\
\cmidrule(lr){2-4} \cmidrule(lr){5-7}
 & \textbf{VG} & \textbf{VC-L} & \textbf{VC-M} & \textbf{VG} & \textbf{VC-L} & \textbf{VC-M} \\
\midrule
Cost (\$) & 0.26 & 3.49 & 3.02 & 0.59 & 4.43 & 6.08 \\
\bottomrule
\end{tabular}
}
\end{table}

Furthermore, under the same \texttt{gpt-5.2} backend, Vibe Graphing reduces API costs by roughly an order of magnitude compared to Vibe Coding, as shown in Table~\ref{tab:vibe_cost}. Workflows generated via Vibe Coding frequently exhibit logical flaws in the constructed graphs, failing to return correct execution results. Consequently, we restrict our comparison with Vibe Coding to cost analysis and exclude it from the performance evaluation. 

To summarize, composed components enhance reusability, while Vibe Graphing significantly lowers development barriers by minimizing boilerplate and manual configuration.
This reduction is particularly important for MAS research prototyping, where developers often need to compare multiple collaboration patterns and repeatedly modify role definitions, communication paths, and context configurations.
By combining declarative specifications with reusable composed components, MASFactory shifts effort from writing orchestration code to designing and refining the collaboration logic itself.
We provide a Vibe Graphing case study in Appendix~\ref{app:vg_case}, which walks through the workflow construction pipeline from natural-language intent to a structured workflow specification and executable workflow definitions.

\section{Conclusion}
We present MASFactory, a graph-centric framework for orchestrating LLM-based multi-agent systems. 
MASFactory models multi-agent workflows as executable directed graphs and integrates Vibe Graphing to compile natural-language intent into MAS workflows with human-in-the-loop refinement.
It also provides reusable components, pluggable context and a visualizer for topology preview, runtime tracing, and visual human-in-the-loop interaction.
Experiments on seven public benchmarks show that MASFactory consistently reproduces representative MASs and that Vibe Graphing produces competitive workflows while substantially reducing implementation overhead.
\section*{Limitations}
MASFactory currently does not provide built-in checkpointing for resuming execution from intermediate states after interruptions. In addition, we will continue to enrich and refine the built-in composed-component library in future updates.

\section*{Acknowledgments}
This work is supported in part by the National Natural Science Foundation of China (No. 62550138, 62192784, 62572064, 62472329), and the Beijing Natural Science Foundation (No. 253004).

\bibliography{custom}
\clearpage

\appendix
\section{Vibe Graphing Case Study}

\label{app:vg_case}

Here we present a concrete Vibe Graphing example that illustrates the end-to-end pipeline from natural-language intent to a structured workflow specification and executable workflow definitions.

\subsection{Intent and Structural Constraint}
We start from a natural-language build instruction that specifies both the task intent (weekly report writing) and an explicit structural constraint:
\emph{START $\rightarrow$ A,B,C $\rightarrow$ D $\rightarrow$ END},
where $A$, $B$, and $C$ are parallel drafting agents and $D$ is an evaluator/selector agent that produces the final output.

The build instruction used in this case study is 
\emph{Design a workflow for writing my weekly report. I will provide what I worked on this week at the beginning. Then run three agents in parallel to draft separate reports, and pass all drafts to a fourth agent to evaluate and select the best one as the final output. The expected workflow structure is: START$\rightarrow$A,B,C$\rightarrow$D$\rightarrow$END.}

\subsection{Interactive Workflow Construction}

\begin{figure}[t]
    \centering
    \includegraphics[width=\linewidth]{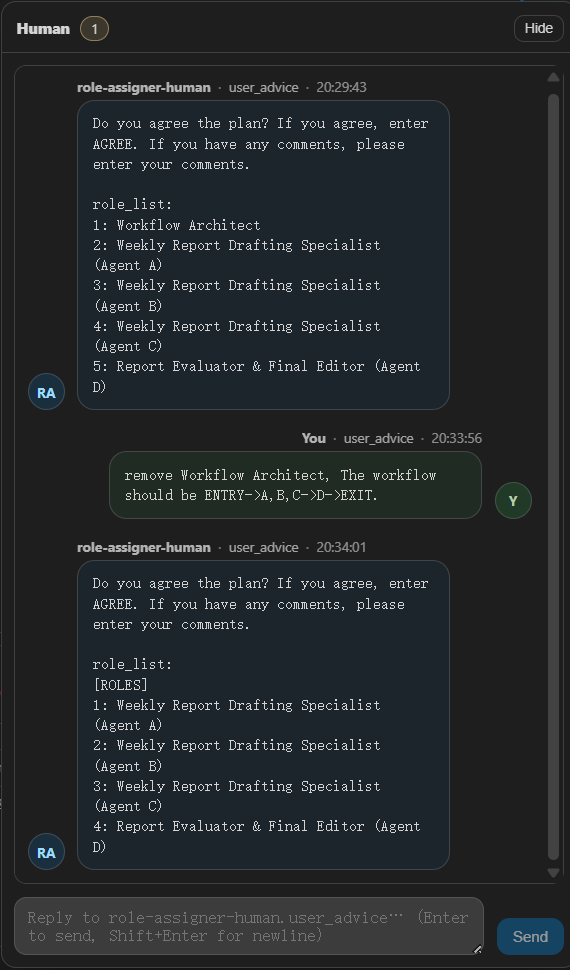}
    \caption{Human-in-the-loop interaction during Vibe Graphing in the visualizer. The user provides feedback to refine the role assignment and structure across stages.}
    \label{fig:vg_human_interact}
\end{figure}
\begin{figure*}[t]
    \centering
    \begin{minipage}[t]{0.49\textwidth}
        \centering
        \includegraphics[width=\linewidth]{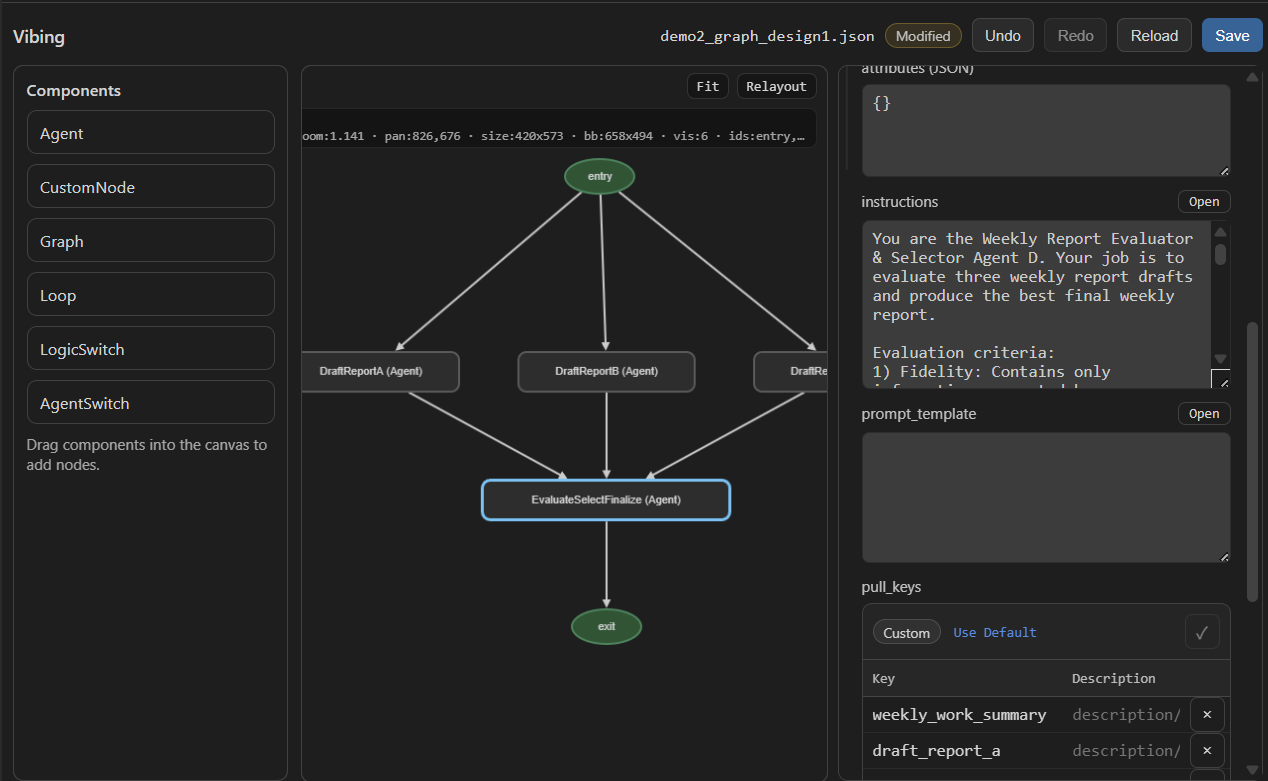}
        \vspace{-0.5em}
        {\small (a) IR preview and editing in the MASFactory Visualizer.}
    \end{minipage}\hfill
    \begin{minipage}[t]{0.49\textwidth}
        \centering
        \includegraphics[width=\linewidth]{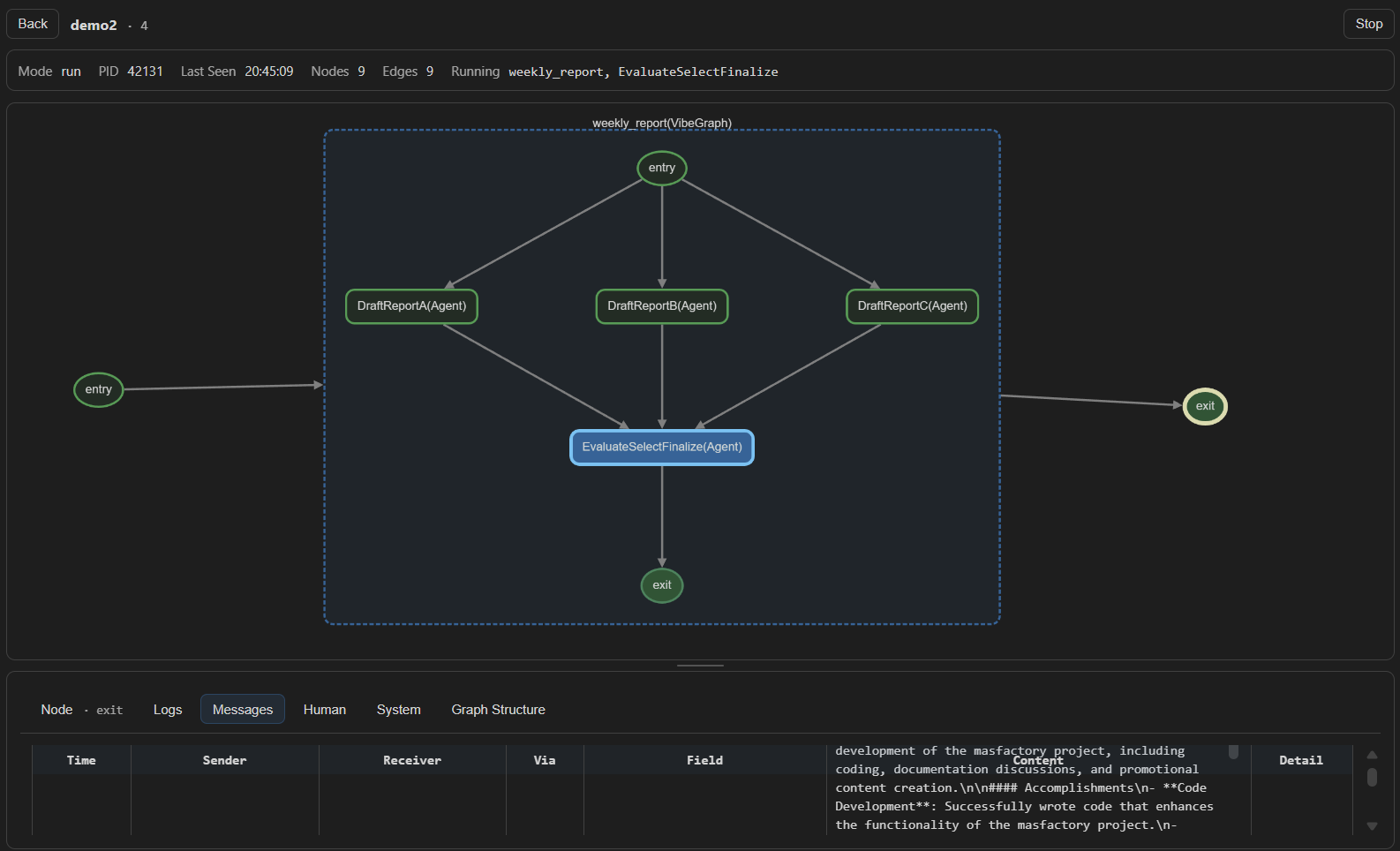}
        \vspace{-0.5em}
        {\small (b) Runtime preview and trace alignment in the MASFactory Visualizer.}
    \end{minipage}

    \caption{Visualizer views used in the Vibe Graphing.}
    \label{fig:viz_two_views}
\end{figure*}

\begin{figure*}[t]
\centering
\fbox{
\begin{minipage}{\textwidth}
\footnotesize\ttfamily
from masfactory import VibeGraph, NodeTemplate, OpenAIModel, RootGraph\\
invoke\_model = OpenAIModel(api\_key=os.environ.get("OPENAI\_API\_KEY",""),\\
\hspace*{10pt}base\_url=os.environ.get("OPENAI\_BASE\_URL",""), model\_name="gpt-4o-mini")\\
build\_model = OpenAIModel(api\_key=os.environ.get("OPENAI\_API\_KEY",""),\\
\hspace*{10pt}base\_url=os.environ.get("OPENAI\_BASE\_URL",""), model\_name="gpt-5.2")\\[4pt]
weekly\_report = NodeTemplate(\\
\hspace*{10pt}VibeGraph,\\
\hspace*{10pt}invoke\_model=invoke\_model,\\
\hspace*{10pt}build\_model=build\_model,\\
\hspace*{10pt}build\_instructions="... START->A,B,C->D->END.",\\
\hspace*{10pt}build\_cache\_path=GRAPH\_DESIGN\_CACHE\_PATH,\\
\hspace*{10pt}pull\_keys=\{"my\_works":"what I worked on this week"\},\\
\hspace*{10pt}push\_keys=\{"final\_weekly\_report":"final weekly report"\},\\
)\\[4pt]
root = RootGraph(name="demo2", nodes=[("weekly\_report", weekly\_report)],\\
\hspace*{10pt}edges=[("ENTRY","weekly\_report",\{\}), ("weekly\_report","EXIT",\{\})])\\
root.build()\\[4pt]
msg, attr = root.invoke(\{\}, \{"my\_works": my\_works\})\\
print(attr["final\_weekly\_report"])\\
\end{minipage}
}
\caption{MASFactory programming code for the Vibe Graphing.}
\label{fig:vg_code}
\end{figure*}

\begin{figure*}[t]
\centering
\fbox{
\begin{minipage}{\textwidth}
\footnotesize\ttfamily
\{\\
\hspace*{6pt}"edges": [\\
\hspace*{10pt}\{"source":"ENTRY","target":"DrafterA"\},\{"source":"ENTRY","target":"DrafterB"\},\\
\hspace*{10pt}\{"source":"ENTRY","target":"DrafterC"\},\{"source":"DrafterA","target":"Finalizer"\},\\
\hspace*{10pt}\{"source":"DrafterB","target":"Finalizer"\},\{"source":"DrafterC","target":"Finalizer"\},\\
\hspace*{10pt}\{"source":"Finalizer","target":"EXIT"\}\\
\hspace*{6pt}],\\
\hspace*{6pt}"nodes": [\\
\hspace*{10pt}\{"id":"DrafterA","type":"Action","input\_fields":["my\_work"],\\
\hspace*{18pt}"output\_fields":["draft\_report\_a"],"instructions":"You are Weekly Report Drafting Agent A ..."\},\\
\hspace*{10pt}\{"id":"DrafterB","type":"Action","input\_fields":["my\_work"],\\
\hspace*{18pt}"output\_fields":["draft\_report\_b"],"instructions":"You are Weekly Report Drafting Agent B ..."\},\\
\hspace*{10pt}\{"id":"DrafterC","type":"Action","input\_fields":["my\_work"],\\
\hspace*{18pt}"output\_fields":["draft\_report\_c"],"instructions":"You are Weekly Report Drafting Agent C ..."\},\\
\hspace*{10pt}\{"id":"Finalizer","type":"Action",\\
\hspace*{18pt}"input\_fields":["my\_work","draft\_report\_a","draft\_report\_b","draft\_report\_c"],\\
\hspace*{18pt}"output\_fields":["final\_weekly\_report","selection\_rationale"],\\
\hspace*{18pt}"instructions":"You are the Weekly Report Evaluator ..."\}\\
\hspace*{6pt}]\\
\}\\
\end{minipage}
}
\caption{Workflow specification produced by Vibe Graphing.}
\label{fig:vg_ir}
\end{figure*}
As shown in Figure~\ref{fig:vg_code}, we write a short program that constructs a minimal wrapper graph with the topology
\texttt{ENTRY} $\rightarrow$ \texttt{VibeGraph} $\rightarrow$ \texttt{EXIT}.
Here, \texttt{VibeGraph} is a built-in composed graph in MASFactory that activates the Vibe Graphing pipeline upon graph construction.
As illustrated in Figure~\ref{fig:vibegraphing}, the pipeline comprises three stages.
Each stage is encapsulated by a \texttt{Loop} component and centers around an \texttt{Agent} node augmented with auxiliary mechanisms for correction, review, and human interaction through \texttt{Interaction}.
After each stage produces an intermediate design, the system solicits user feedback and iteratively refines the result until the user accepts it, at which point the pipeline proceeds to the next stage.
During the human-in-the-loop interactions in Stage~2 and Stage~3, users may either (i) edit the structured intermediate representation directly via the visualizer, or (ii) provide feedback in the interaction panel.
Both manual edits and textual feedback are recorded and fed back to the \texttt{Agent} as references for subsequent revisions.
Figure~\ref{fig:vg_human_interact} shows an example interaction trace in the visualizer, and Figure~\ref{fig:viz_two_views}(a) shows the workflow preview and editing interface.

\subsection{Workflow Specification Snapshot}
As shown in Figure~\ref{fig:vg_ir}, we provide a snapshot of the structured intermediate representation generated by Vibe Graphing.
The representation explicitly encodes: (i) node semantics (e.g., \texttt{label} and \texttt{instructions}), (ii) input/output contracts (e.g., \texttt{input\_fields} and \texttt{output\_fields}), and (iii) directed dependencies among nodes (i.e., \texttt{edges}).

\subsection{Workflow Execution}
At runtime, the visualizer provides a synchronized view of execution states and message traces, as shown in Figure~\ref{fig:viz_two_views}(b).
The workflow first collects user input and normalizes it into \texttt{weekly\_work\_summary}, then executes three drafting branches in parallel, and finally aggregates drafts for selection and light editing.

\end{document}